%% file: main.tex
\documentclass[10pt, conference, compsocconf]{IEEEtran}      
\def\IEEEbibitemsep{0pt plus .5pt}
\usepackage{graphicx}
\usepackage{amsmath}
\usepackage{amsfonts}
\usepackage[bookmarks=false,pdfborder={0 0 0}]{hyperref}
\graphicspath{{./Images/}{./Cherries_casme/}{./Cherries_cross/}{./Bad_Cherries/}}

\begin{document}

\IEEEoverridecommandlockouts\pubid{\makebox[\columnwidth]{978-1-5386-2335-0/18/\$31.00~\copyright{}2018 IEEE \hfill}
\hspace{\columnsep}\makebox[\columnwidth]{ }}

\title{\LARGE
Enriched Long-term Recurrent Convolutional Network\\for Facial Micro-Expression Recognition
}%
\author{\parbox{16cm}{\centering
    {\large Huai-Qian Khor$^1$, John See$^2$, Raphael C.W. Phan$^3$, Weiyao Lin$^4$}\\
    {\normalsize
    $^{1,}$$^2$ Faculty of Computing and Informatics, Multimedia University, Malaysia\\
    $^3$ Faculty of Engineering, Multimedia University, Malaysia\\
    $^4$ Department of Electronic Engineering, Shanghai Jiao Tong University, China\\
    Emails: $^1$hqkhor95@gmail.com, $^2$johnsee@mmu.edu.my, $^3$raphael@mmu.edu.my, $^4$wylin@sjtu.edu.cn}}
}

\maketitle

\begin{abstract}
Facial micro-expression (ME) recognition has posed a huge challenge to researchers for its subtlety in motion and limited databases. Recently, handcrafted techniques have achieved superior performance in micro-expression recognition but at the cost of domain specificity and cumbersome parametric tunings. In this paper, we propose an Enriched Long-term Recurrent Convolutional Network (ELRCN) that first encodes each micro-expression frame into a feature vector through CNN module(s), then predicts the micro-expression by passing the feature vector through a Long Short-term Memory (LSTM) module. The framework contains two different network variants: (1) Channel-wise stacking of input data for spatial enrichment, (2) Feature-wise stacking of features for temporal enrichment. We demonstrate that the proposed approach is able to achieve reasonably good performance, without data augmentation. In addition, we also present ablation studies conducted on the framework and visualizations of what CNN "sees" when predicting the micro-expression classes.   

\end{abstract}


\input{Introduction}

\input{RelatedWorks}
\input{Framework}

\input{Experiment}

\input{Discussion}
\input{Conclusion}
\section*{Acknowledgments}
This work was supported in part by MOHE Grant FRGS/1/2016/ICT02/MMU/02/2 Malaysia and Shanghai 'The Belt and Road' Young Scholar Exchange Grant (17510740100). The authors would like to thank the anonymous reviewers for their helpful and constructive comments. We are also grateful to our lab colleagues for sharing computational resources.





\bibliographystyle{IEEEtran}
\bibliography{references}

\end{document}

%% file: Introduction.tex
\section{Introduction}

Facial micro-expressions (ME) are brief and involuntary rapid facial emotions that are elicited to hide a certain true emotion \cite{ekman_nonverbal}. A standard micro-expression lasts between 1/5 to 1/25 of a second and usually occurs in only specific parts of the face \cite{ekman_constant}. The subtleness and brevity of micro-expressions are a great challenge to the naked eye; hence, a lot of works have been proposed in recent years to utilize computer vision and machine learning algorithms in attempt to achieve automated micro-expression recognition.

The establishment of Facial Action Coding System (FACS) \cite{ekman_facial} encodes the facial muscle changes to emotion states. The system also establishes a ground truth of the exact begin and end time of each action unit (AU). Different databases \cite{smic, casme2, samm} may contain different micro-expression classes which are labeled by trained coders based on the presence of AUs. 
However, a recent discourse by Davison et al. \cite{davison_objective} argued that using AUs instead of emotion labels can define micro-expressions more precisely since the training process can learn based on specific facial muscle movement patterns. They further proved that this leads to higher classification accuracy.

In this field of research, several works \cite{liong_lessismore}\cite{huang_stclqp}\cite{huang_ip} have achieved impressive micro-expression recognition performance. These works have proposed carefully crafted descriptors and/or methods that involved a tedious tuning of parameters to attain maximum results. In view of these unwieldy steps, the adoption of \emph{deep learning} techniques or deep neural networks have started to take-off, as seen from several new attempts \cite{kim_state,peng_dual}. However, the usage of deep neural network poses challenges to ME recognition due to the scarcity of samples and class-imbalance in most micro-expression data. 

%% file: RelatedWorks.tex
\section{Related Works}

\subsection{Handcrafted Features}

In the last five years, numerous works have been proposed to solve the ME recognition problem. 
The databases established to advance computational research in spontaneous facial micro-expression analysis i.e. SMIC \cite{smic}, CASME II \cite{casme2}, SAMM \cite{samm,davison_objective}, have mainly chosen Local Binary Pattern with Three Orthogonal Planes (LBP-TOP) \cite{zhao_lbptop} as their primary baseline feature extractor. The LBP-TOP is a spatio-temporal extension of the classic Local Binary Pattern (LBP) descriptor \cite{ojala_lbp}, which characterizes the local textural information by encoding a vector of binary code into histograms. LBP-TOP extracts the said histograms from each of the three planes (XY, XT, YT) and concatenate them into a single feature histogram.  
The LBP, whilst known for its simplicity in computation, is vastly used 
because of its robustness towards illumination changes and image transformations. 

Wang et al. \cite{wang_lbpsip} reduced the redundancies in the LBP-TOP by utilizing only six intersection points in the 3D plane to construct the feature descriptor. Later on, Huang et al. \cite{huang_ip} proposed a Spatio-Temporal LBP with Integral Projection (STLBP-IP) that applies the LBP operator on horizontal and vertical projections based on difference images. Their method is shape-preserving and is robust against the white noise and image transformations.  

Several works have used LBP-TOP with an accompanying pre-processing technique. Most widely seen is the Temporal Interpolation Model \cite{smic} which is used to sample uniformly a fixed number of image frames from the constructed data manifold. Recently, \cite{cat_dmdsp} proposed Sparsity Promoting Dynamic Mode Decomposition (DMDSP) which acts to select only the significant temporal dynamics when synthesizing a dynamically condensed sequence. A number of other works \cite{li2017towards,wang2017effective} opt to magnify the video in attempt to accentuate the subtle changes before feature extraction.

Motion information can readily portray the subtle changes exhibited by micro-expressions. Shreve et al. \cite{shreve_strain} proposed the extraction of a derivative of optical flow called an \emph{optical strain} which was originally used for ME spotting but later adopted as a feature descriptor for ME recognition \cite{liong_subtle,liong_spontaneous}. 
Leveraging on the discriminativeness of optical flow, other interesting approaches have come to the fore, among them are Bi-Weighted Oriented Optical Flow (Bi-WOOF) \cite{liong_lessismore} and Facial Dynamics Map \cite{xu_facialmaps}.

\subsection{Deep Neural Networks}
The utilization of deep learning techniques or deep neural networks is fairly new to this field of research despite its widespread popularity in recognition tasks.

One early work \cite{kim_state} to utilize deep learning proposed an expression-state based feature representation. The researchers adopted Convolutional Neural Networks (CNN) to encode different expression states (i.e., onset, onset to apex, apex, apex to offset and offset). Several objective functions are optimized during spatial learning to improve expression class separability. After that, the encoded features are passed to a Long Short-Term Memory (LSTM) network to learn time scale dependent features.  

Recently, Peng et al. \cite{peng_dual} proposed a two-stream 3-D CNN model called Dual Temporal Scale Convolutional Neural Network (DTSCNN). Different streams of the framework were used to adapt to different frame rates of ME video clips. The authors aggregated both CASME I and II databases, likely to provide sufficient samples for meaningful training to take place. The network was also designed to be shallower to avoid overfitting problem, while optical flow was used to enrich the input data. These two approaches provide the motivation towards the design of our proposed method.

%% file: Framework.tex
\section{Proposed Framework}


\begin{figure*}[!t]
\centering
\includegraphics[scale=0.4]{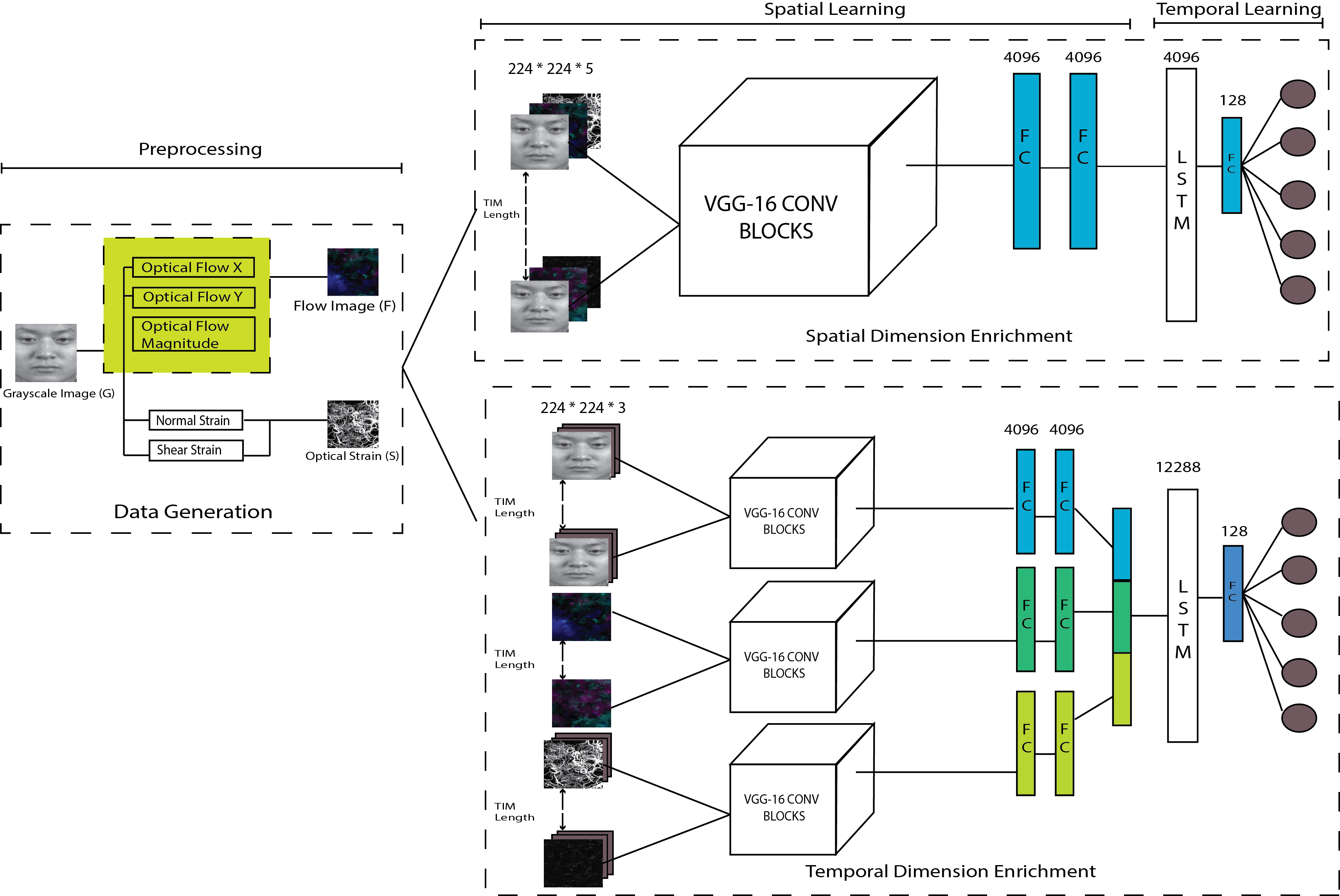}
\caption{Proposed ELRCN framework}
\label{fig:lrcn}
\end{figure*}

In this work, we propose an Enriched Long-term Recurrent Convolutional Network (ELRCN) for micro-expression recognition, which adopts the architecture of \cite{lrcn} whilst performing feature enrichment to encode subtle facial changes. The ELRCN model comprises of a deep hierarchical spatial feature extractor and a temporal module that characterizes temporal dynamics. Two variants of the network are introduced: 1) Enrichment of the spatial dimension by input channel stacking, 2) Enrichment of the temporal dimension by deep feature stacking. Figure \ref{fig:lrcn} summarizes the proposed framework with the preprocessing module and both variants of learning module.

\subsection{Preprocessing}
The micro-expression videos are first preprocessed using TV-L1 \cite{zach_tvl1flow} method for optical flow approximation, which has two major advantages: better noise robustness and preservation of flow discontinuities. Optical flow encodes motion of an object in vectorized notations, indicating the direction and intensity of the motion or `flow' of image pixels. The horizontal and vertical components of the optical flow are defined as follow:

\begin{equation}
	\vec{v} = [p = \frac{dx}{dt}, q = \frac{dy}{dt}] ^T
    \label{optical_flow}
\end{equation}
where $dx$ and $dy$ represent the estimated changes in pixels along the $x$ and $y$ dimension respectively while \textit{dt} represent the change in time. To form a 3-dimensional \emph{flow image}, we concatenate the horizontal and vertical flow images, \textbf{p} and \textbf{q} and the optical flow magnitude, $\textbf{m} = |v|$.
Normalization of the flow image is not necessary 
in our case since motion changes are very subtle (not occupying large range of values); this was also proven empirically with negligible drop in performance. 

We also obtained the optical strain \cite{shreve_strain} by computing the derivatives of the optical flow. By employing optical strain, we are able to properly characterize the tiny amount of movement of a deformable object present between two successive frames. This is described by a displacement vector, $\textbf{u} = [u, v]^T$. The finite strain tensor is defined as:

\begin{equation}
	\epsilon = \frac{1}{2}[\nabla u + (\nabla u)^T]
    \label{epsilon}
\end{equation}
or in expanded tensor form:

\begin{equation}
	\epsilon =  \begin{bmatrix}
	\epsilon_{xx} = \frac{\delta u}{\delta x} & \epsilon_{xy} = \frac{1}{2}(\frac{\delta u}{\delta y} + \frac{\delta v}{\delta x}) \\
    \epsilon_{yx} = \frac{1}{2}(\frac{\delta v}{\delta x} + \frac{\delta u}{\delta y}) & \epsilon_{yy} = \frac{\delta v}{\delta y}
	\end{bmatrix} 
    \label{bimatrix}
\end{equation}
where the diagonal strain components, ($\epsilon_{xx}$, $\epsilon_{yy}$), are normal strain components and ($\epsilon_{xy}$, $\epsilon_{yx}$) are the shear strain components. Normal strain measures changes along $x$ and $y$ directions whereas shear strain measures changes in the angles caused by deformation along both axis. 












The optical strain magnitude for each pixel can be computed using the sum of squares of the normal and shear strain components:

\begin{equation}
	|\epsilon| = \sqrt[]{\epsilon_{xx}^2 + \epsilon_{yy}^2 + \epsilon_{xy}^2 + \epsilon_{yx}^2}
\end{equation}

\subsection{Spatial Learning}
Recent deep models \cite{alex_imagenet,krizhesky_imagenet,he_resnet,szegedy_deeper} have proven that the composition of numerous ``layers'' of non-linear functions can achieve ground-breaking results for various computer vision problems such as object recognition and object detection. To leverage the benefit of deep convolutional neural networks (CNN) in a sequential fashion, the input data $x$ is first encoded with a CNN to a fixed-length vector, $\phi (x_t)$ that represents the spatial features at time $t$. Subsequently, $\phi (x_t)$ is then passed to a recurrent neural network to learn the temporal dynamics.

In this paper, we also hypothesize that by using additional derivative information of the raw input sample, in a process that involves \textit{sample enrichment}, we can minimize under-fitting in the learned models, which in turn can result in higher recognition performance. Figure \ref{fig:lrcn} depicts the overall framework of our proposed Enriched Long-term Recurrent Convolutional Network (ELRCN) in two possible variants: Spatial Dimension Enrichment (SE) and Temporal Dimension Enrichment (TE). 

The SE model uses a larger input data dimension for spatial learning by stacking an optical flow image ($F\in\mathbb{R}^3$), an optical strain image ($S \in \mathbb{R}^2$) and a gray-scale raw image ($R \in \mathbb{R}^2$) along the input channel, which we denote as 
$x_t = (F_t, S_t, G_t)$. Hence, the input data is 224 * 224 * 5, which necessitates training the VGG-Very-Deep-16 (VGG-16) \cite{simonyan_verydeep} model from scratch. The last fully connected (FC) layer encodes the input data into a 4096 fixed-length vector, $\phi (x_t)$.

The TE model utilizes transfer learning \cite{bengio_transfer} with pre-trained weights from VGG-Face model \cite{parkhi_vggfaces} which was trained on a large-scale Labeled Faces in the Wild (LFW) dataset \cite{huang_lfw} for the purpose of face identification. 
We 
fine-tuned the micro-expression data on the pre-trained weights of VGG-Face to allow the model to learn and adapt more effectively. This also facilitates faster convergence because both micro-expression and LFW data involve faces and their components. 
Since the VGG-Faces model expects a 224 * 224 * 3 input, we duplicated the S and G images ($\mathbb{R}^2\rightarrow\mathbb{R}^3$) so that they fit the required input dimension 
(as shown in Figure \ref{fig:lrcn}). During the training phase, we fine-tune each input data in separate VGG-16 models with each model yielding a 4096-length feature vector, $\phi (x_t)$ at their last FC layer. This results in 
a 12288-length feature vector to be passed to the subsequent recurrent network.

\subsection{Temporal Learning}
In the current micro-expression domain, several works \cite{zheng_rksvd,liong_spontaneous} aimed to preserve the temporal dimension as its dynamics is crucial for recognizing facial movements. We use a popular variant of the recurrent neural network called Long Short-Term Memory (LSTM) \cite{gers_lstm} to learn the spatially-encoded sequential input, $\phi (x_t)$. The LSTM seeks to learn weights parameters $W$ that maps the input $\phi(x_t)$ at a previous time step hidden state $h_{t-1}$ to an output $z_t$ and updated hidden state $h_t$. LSTM layers can be stacked serially, followed by a fully connected layer that encodes $z_t$ into a smaller dimension, $\hat{y} = W_zz_t + b_z$. Finally, the prediction $P(y_t)$ is computed with softmax of $\hat{y}_t$:
\begin{equation}
	P(y_t = c) = \text{softmax}(\hat{y}_t) = \frac{exp(\hat{y}_t, c)}{\sum\limits_{c' \in C} exp(\hat{y}_t, c')}
    \label{eq:lstm_softmax}
\end{equation}    
where C is a discrete, finite set of outcomes and $y_t \in C$.

\subsection{General Network Configuration}
The networks are trained using adaptive epochs or early stopping with a maximum set to 100 epochs. Basically, the training for each fold will stop when the loss score stops improving. We use Adaptive Moment Estimation (ADAM) \cite{king_adam} as the optimizer, with a learning rate of $10^{-5}$ and decay of $10^{-6}$. The learning rate is tuned to be smaller than typical rates because of the subtleness of micro-expression which poses difficulty for learning. 
For temporal learning, we fix the number of FC layers after the LSTM layers to one. This is not experimented as our focus is on the number of recurrent layers and units in these layers (see the ablation study in Section \ref{sec:ablation}).

%% file: Experiment.tex
\section{Evaluation}

\subsection{Databases}
CASME II \cite{casme2} is a comprehensive spontaneous micro-expression database containing 247 video samples, elicited from 26 Asian participants with an average age of 22.03 years
old. The videos in this database showed a participant evoked by one of five categories of micro-expressions: Happiness, Disgust, Repression, Surprise, Others.

The Spontaneous Actions and Micro-Movements (SAMM) \cite{samm} is a newer database of 159 micro-movements (one video for each) induced spontaneously from a demographically diverse group of 32 participants with a mean age of 33.24 years, and an even male-female gender split. Originally intended for investigating micro-facial movements, the SAMM was induced based on the 7 basic emotions. Eventually, the authors \cite{davison_objective} proposed ``objective classes'' based on the FACS Action Units as categories for micro-expression recognition. 

Both the CASME II and SAMM databases have much in common: They are recorded at a high speed frame rate of 200 \emph{fps}, and they have objective classes, as provided in \cite{davison_objective}.

\subsection{Preprocessing \& Settings}

The SAMM dataset is preprocessed with Dlib \cite{dlib_king} for face alignment while facial landmarks are extracted using Face++ API \cite{faceplusplus}. Then, each video frame is cropped based on selected facial landmarks at the edge of the face. Meanwhile, CASME II provides pre-cropped video frames which we make use directly. All video frames are resized to $224 * 224$ pixel resolution to match the input spatial dimension to the network. Temporal Interpolation Model (TIM) \cite{smic} of length 10 was applied to both databases to fit the sample sequence into the recurrent model that expects a fixed temporal length. The baseline methods that we compared with were implemented using a Support Vector Machine (SVM) 
with linear kernel and a large regularization parameter of C=10000. 

We perform two sets of experiments: (1) Single domain experiment involving only one database (CASME II), (2) Cross domain experiment involving two databases (CASME II and SAMM), specifically, two settings were used -- one which holds out one database at each time, another which combines all samples from both databases. 

Experiments are measured using F1-Score, Weighted Average Recall (WAR) or Accuracy, and Unweighted Average Recall (UAR). UAR is akin to a ``balanced" accuracy (averaging the accuracy scores of each individual class without consideration of class size). We report micro-averaged F1-Score, which provides a balanced metric when considering highly imbalanced data \cite{le2014imbalanced}.



\subsection{Single Domain Experiment}
In this experiment, the CASME II database is our choice of domain for evaluation. Training was performed using Leave-One-Subject-Out (LOSO) cross validation as this protocol prevents subject bias during learning. 
Table \ref{table:singledb} compares the performance of our proposed methods against the baseline LBP-TOP method (reproduced) and a number of recent and relevant works in literature. The TE variant of the proposed ELRCN method clearly outperforms its SE counterpart, which shows the importance of fine-tuning separate networks for each type of data.

\begin{table}[ht]
\centering
\caption{Performance of proposed methods vs. other methods for micro-expression recognition}
\begin{tabular}{ | l | c | c | c |}
\hline

Methods & F1-Score & UAR & Accuracy / WAR \\ 
\hline
LBP-TOP (reproduced) & 0.2941 & 0.3094 & 0.4595 \\ 
\hline
Adaptive MM + & N/A & N/A & 0.5191 \\
LBP-TOP \cite{park_subtle} & & & \\
\hline
FDM \cite{xu_facialmaps} & 0.4053 & N/A & 0.4593 \\ 
\hline
LBP-SIP & 0.4480 & N/A & 0.4656 \\
\hline
EVM+HIGO \cite{li2017towards} & N/A & N/A & \textbf{0.6721} \\
\hline
CNN-LSTM \cite{kim_state} & N/A & N/A & 0.6098\\
\hline
ELRCN-SE & 0.4547 & 0.3895 & 0.4715 \\
\hline
ELRCN-TE & \textbf{0.5000} & \textbf{0.4396} & 0.5244 \\ 
\hline
\end{tabular}
\label{table:singledb}
\end{table}

\begin{table}[t!]
\centering
\caption{Experimental results for CDE evaluation}

\begin{tabular}{ | l | c | c | c |}
\hline 
Methods & F1-Score & UAR & Accuracy / WAR \\
\hline
LBP-TOP (reproduced) & 0.3172 & 0.3224 & 0.4229 \\
\hline
ELRCN-SE & \textbf{0.4107} & \textbf{0.3900}  & \textbf{0.5700} \\
\hline
ELRCN-TE & 0.3616 & 0.3300 & 0.4700 \\
\hline
\end{tabular}
\label{table:composite}
\end{table}

\begin{figure}[!hbt]
\includegraphics[width=0.45\textwidth]{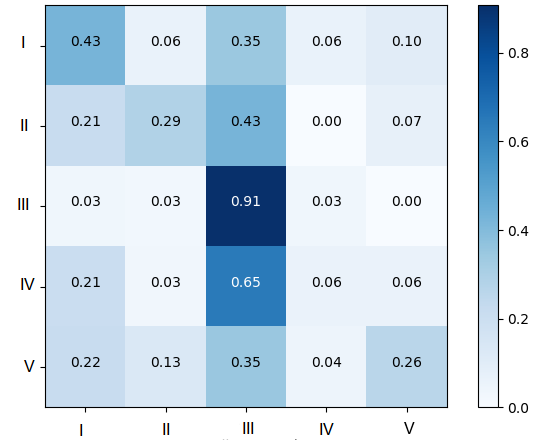}
\vspace{-0.5em}
\caption{Confusion Matrix of ELRCN-SE on CDE protocol.}
\label{fig:cm_cde}
\vspace{0em}
\end{figure} 

\subsection{Cross Domain Experiment}
To test the robustness of our deep neural network architecture and its ability to learn salient characteristics from the samples, we use two cross domain protocols introduced by the Micro-Expression Grand Challenge (MEGC) 2018\footnote{\url{http://www2.docm.mmu.ac.uk/STAFF/m.yap/FG2018Workshop.htm}} 
-- Composite Database Evaluation (CDE) and Holdout-Database Evaluation (HDE). HDE and CDE are Tasks A and B respectively in MEGC 2018. CDE combines both databases (CASME II and SAMM), which totals to 47 subjects after omitting the 6th and 7th objective classes (from \cite{davison_objective}) followed by a LOSO evaluation. HDE samples the training and test sets from opposing database (i.e, trains on CASME II and test on SAMM, and vice versa). The results from both folds are then averaged and reported as the overall result. 


Table \ref{table:composite} compares the performance of our two ELRCN variants against the reproduced LBP-TOP baseline on the CDE (Task B) protocol. The proposed methods are clearly superior in generalizing over a large number of subjects as compared to the baseline method. Interestingly, the SE variant posts a much stronger result (WAR 0.57) than the TE variant; this in contrast to results on CASME II alone. 

Table \ref{table:holdout} shows the result for the HDE (Task A) protocol. The HOG-3D and HOOF methods were provided by the challenge organizers as other competing baselines. We also reproduced the baseline LBP-TOP method which differed from the results provided by the challenge organizers. This is likely to due to some differences in the face cropping steps or preprocessing steps (such as TIM) which were not disclosed in detail at the time of writing.
Similarly, we observe a strong performance from the SE variant of the proposed approach, which surpasses that of the TE variant and the provided baselines.

\begin{figure}[!t]
\includegraphics[width=0.4\textwidth]{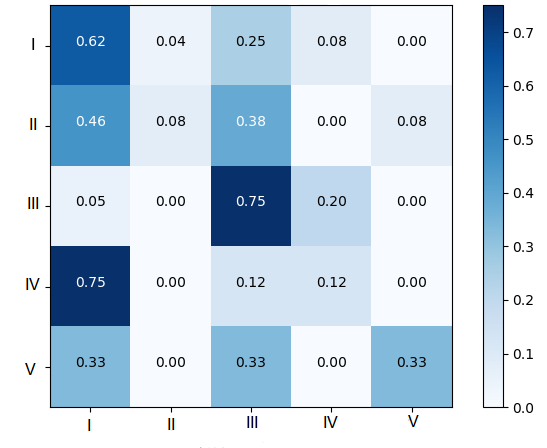}
\vspace{-0.5em}
\caption{Confusion matrix of ELRCN-SE on HDE protocol, training on CASME II and testing on SAMM database.}
\label{fig:cm_casme_samm}
\vspace{0em}
\end{figure}

\begin{figure}[!hbt]
\includegraphics[width=0.4\textwidth]{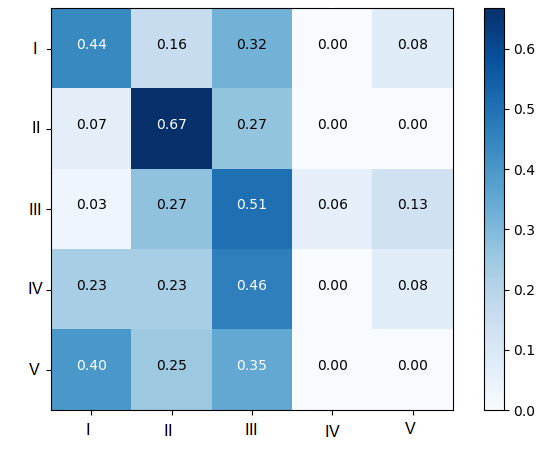}
\vspace{-0.5em}
\caption{Confusion matrix of ELRCN-SE on HDE protocol, training on SAMM  and testing on CASME II database.}
\label{fig:cm_samm_casme}
\vspace{0em}
\end{figure}

To better understand what goes on under the hood, we provide the confusion matrix for ELRCN-SE with CDE protocol in \ref{fig:cm_cde}. Class I and class III have the best results possibly due to larger amount of training samples. Besides, we also provide the confusion matrices for both folds (i.e. train-test pairings of CASME II-SAMM and SAMM-CASME II) in Figures \ref{fig:cm_casme_samm} and \ref{fig:cm_samm_casme}. The CASME II-SAMM fold (F1 0.409, UAR 0.485, WAR 0.382) had noticeably better performance than the SAMM-CASME II fold (F1 0.274, UAR 0.384, WAR 0.322). Class III of CASME II has the most training samples; it performed the best. Likewise, classes that were relatively under-represented in the training set (class II from CASME II, classes IV and V from SAMM) performed very poorly. Hence, it is likely that the small sample size remains a stumbling block for deep learning based approaches. 

\begin{table}[t!]
\centering
\caption{Experimental results for HDE evaluation}
\begin{tabular}{ | l | c | c | c |}
\hline
Methods & F1-Score & UAR & Accuracy / WAR \\ \hline
LBP-TOP (reproduced) & 0.2162 & 0.2179 & 0.3891 \\
\hline
LBP-TOP (provided) & N/A & 0.322 & 0.285 \\
\hline
HOG-3D & N/A & 0.228  & 0.363\\
\hline
HOOF & N/A & 0.348 & 0.353 \\
\hline
ELRCN-SE & \textbf{0.3411} & \textbf{0.3522} & \textbf{0.4345} \\
\hline
ELRCN-TE & 0.2389 & 0.2221 & 0.2320 \\
\hline
\end{tabular}
\label{table:holdout}
\end{table}

\subsection{Ablation Study}
\label{sec:ablation}

For further analysis, we perform an extensive ablation study by removing certain portions of our proposed ELRCN to see how that affects performance. This was carried out using the CASME II database (single domain).

\subsubsection{Spatial Learning Only}
\begin{figure}[!t]
\includegraphics[width=0.5\textwidth]{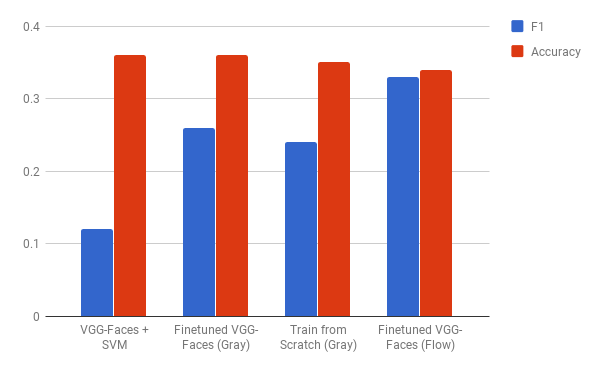}
\vspace{-0.5em}
\caption{Recognition performance using spatial module only}
\label{fig:spatialonly}
\vspace{0em}
\end{figure}

We learn only with the VGG-16 CNN to observe the capability of the spatial module on its own. We regard each video frame as individual images instead of a sequence. Results in Figure \ref{fig:spatialonly} on different configurations of the spatial module show that spatial-only performances can be poorer than that of the baseline. 

\subsubsection{Temporal Learning Only}
\begin{figure}[!hbt]
\includegraphics[width=0.5\textwidth]{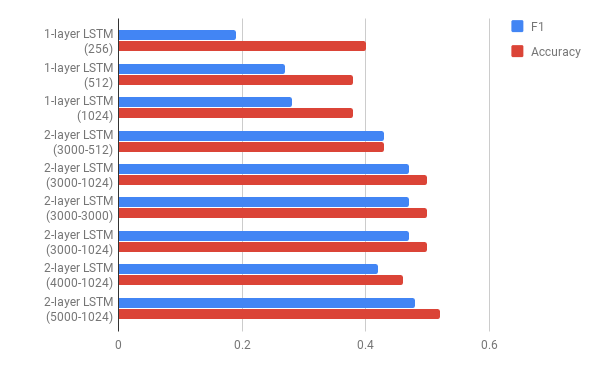}
\vspace{-0.5em}
\caption{Recognition performance using temporal-only module. The numbers in bracket indicate the number of recurrent units for each LSTM layer.}
\label{fig:temporalonly}
\vspace{0em}
\end{figure}

The images are resized to 50*50 pixel resolution since recurrent models with large number of recurrent units are computationally demanding. We consider the pixel intensities as the basic representation of the samples as input to the temporal module. A variety of configurations were considered, including both 1 and 2-layer LSTMs. Results in Figure \ref{fig:temporalonly} show that the baseline performance can be surpassed by just using pixel intensities as input to 2-layer LSTM networks. With reference to the spatial-only approaches, the importance of temporal dynamics is quite telling, as can be seen here.

\subsubsection{Spatio-Temporal LRCN}
From the first two studies, we proceed to gauge the performance of the proposed method (SE variant) by fixing one of the two modules to a reasonably good choice of method and varying the other. 

\begin{figure}[!t]
\includegraphics[width=0.5\textwidth]{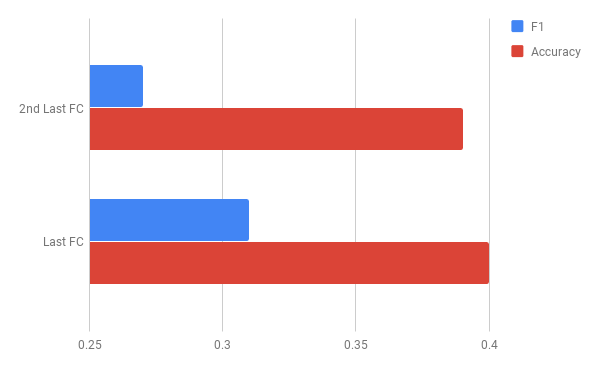}
\vspace{-1em}
\caption{Recognition performance using Flow features encoded at different FC layers (spatial module), on a 2-layer LSTM (3000-1024).}
\label{fig:compare_spat}
\vspace{0em}
\end{figure}

\begin{figure}[!t]
\includegraphics[width=0.5\textwidth]{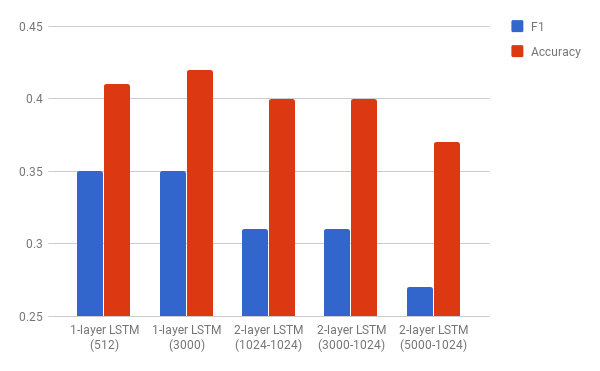}
\vspace{-0.5em}
\caption{Recognition performance of different temporal recurrent networks using spatial features from the last FC of the VGG-16 CNN.}
\label{fig:compare_temporal}
\vspace{0em}
\end{figure}

Using Flow data only (the best from spatial-only study), we tested using spatial features from the last and second last fully-connected (FC) layers of the VGG-16 CNN on a 2-layer LSTM (3000-1024), which is the best architecture as of temporal-only study (see Fig. \ref{fig:temporalonly}). Results in \ref{fig:compare_spat} show that the spatial features are the most discriminative when taken from the 4096-length last FC layer. Following this, the opposing study proceeds to test this selected spatial feature against a number of temporal network architectures. Results in Figure \ref{fig:compare_temporal} shows an interesting case of a single layer LSTM performing better than 2-layer LSTMs in an ELRCN framework, when image-based features are used instead of pixel intensities. 
Additionally, we note that using more recurrent units also do not necessarily produce better results, but with the sure certainty of an increase in computational cost. 

These studies reveal that both spatial and temporal modules have different roles to play within the framework, and they are highly dependent on each other to attain a good level of performance.  


%% file: Discussion.tex
\section{Discussion}

\paragraph{Using more data} The limitations of deep learning techniques is most obvious in the aspect of sample size. Typical deep architectures require a large amount of data to learn well. We experimented with the use of more interpolated frames (higher TIM), but it resulted in poorer results than what was recommended by earlier works \cite{smic,le2014imbalanced}, i.e. TIM of 10 or 15. However, we do expect some improvement if appropriate data augmentation is used on our proposed network.  

\begin{figure*}[t]
\centering
\includegraphics[width=0.14\textwidth]{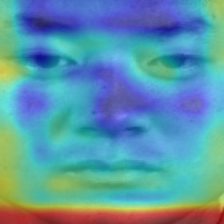}
\includegraphics[width=0.14\textwidth]{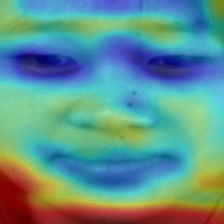}
\includegraphics[width=0.14\textwidth]{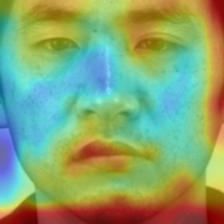}
\includegraphics[width=0.14\textwidth]{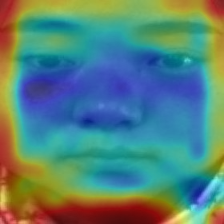}
\includegraphics[width=0.274\textwidth]{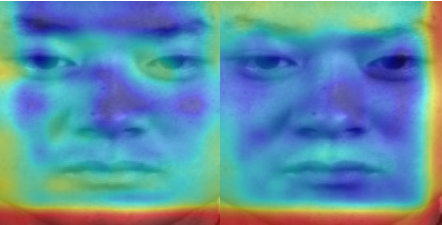}\\
\scriptsize\hspace{-4.25em}(a)\hspace{10em}(b)\hspace{10em}(c)\hspace{10em}(d)\hspace{16em}(e)

\caption{Grad-CAM visualizations of the ELRCN-TE model: \emph{from left:} (a) Subject 1 (CASME II), Happiness; (b) Subject 2 (CASME II), Others; (c) Subject 13 (SAMM), Class III; (d) Subject 5 (CASME II), Objective class III, CDE protocol (e) Subject 1 (CASME II), Happiness, Comparison between single domain and cross domain experiments.}
\label{fig:casme_cam_0}
\vspace{-1em}
\end{figure*}

\paragraph{Visualizations} To better ``see'' how the proposed network arrived at its predictions, we utilize the gradient-weighted class activation mapping (Grad-CAM) \cite{gradcam} on the last convolutional layer of the spatial network to provide visual explanations as to which parts of the face are contributing towards the classification decision. The visualizations in Figure \ref{fig:casme_cam_0} are colored based on colors from the visible light spectrum, ranging from blue (not activated) to red (highly activated). The activations correspond to spatial locations that contribute most to the predicted class. 

We first show the visualizations from the single domain experiment. The AU 12 (lip corner puller) from the sample in Figure \ref{fig:casme_cam_0}(a) correspond quite precisely with the greenish regions near the side of lips. The area around the cheeks of the subject in Figure \ref{fig:casme_cam_0}(b) also show relatively strong activations which corresponds to AU 14, the ground truth. 

From the cross domain experiment, we also found similar evidence of AU-matching spatial activations from Figures \ref{fig:casme_cam_0}(c) and (d). The AUs for Figure \ref{fig:casme_cam_0}(c) are 4, 6, 7, 23, which involves movements around eye regions and upper cheek, both of which are reddish strong. Meanwhile, the sample in Figure \ref{fig:casme_cam_0}(d) has AU 1 that involves raising eyebrows. 
Comparing the Grad-CAMs of a same sample on different experiments (shown in Figure \ref{fig:casme_cam_0}(e)) generally indicate that models trained on a single domain had more salient locations than that on cross domain. 










%% file: Conclusion.tex
\section{Conclusion}
In this paper, we have proposed two variants of an Enriched LRCN model for micro-expression recognition -- one which stacks various input data for spatial enrichment (SE), another which stacks features for temporal enrichment (TE). Empirically, the TE model performs better on a single database while the SE model learns better in cross domain. The Grad-CAM visualization on selected samples demonstrate that the predictions from these models somewhat conform to the AUs marked by experts. Through our ablation study, we also discover that using optical flow information is more beneficial than using raw pixel intensities in providing proper characterization of the input data to the network. In future, we hope to extend our preliminary work with appropriate data augmentation and preprocessing techniques.
